\documentclass{article}

\usepackage[preprint]{neurips_2024}
\usepackage{graphicx}
\usepackage{siunitx} % For aligning numbers by decimal point
\usepackage{makecell} % For multi-line table headers
\usepackage{comment}
\usepackage{amsmath}
\usepackage[utf8]{inputenc} % allow utf-8 input
\usepackage[T1]{fontenc}    % use 8-bit T1 fonts
\usepackage{hyperref}       % hyperlinks
\usepackage{url}            % simple URL typesetting
\usepackage{booktabs}       % professional-quality tables
\usepackage{amsfonts}       % blackboard math symbols
\usepackage{nicefrac}       % compact symbols for 1/2, etc.
\usepackage{microtype}      % microtypography
\usepackage{xcolor}         % colors
\usepackage{todonotes}

\title{Reducing Transformer Key-Value Cache Size with Cross-Layer Attention}

\author{%
  William Brandon\thanks{Equal Contribution} \\
  MIT CSAIL\\
  \texttt{wbrandon@csail.mit.edu} \\
  \And
  Mayank Mishra$^*$ \\
  MIT-IBM Watson AI Lab \\
  \And
  Aniruddha Nrusimha \\
  MIT CSAIL \\
  \And
  Rameswar Panda \\
  MIT-IBM Watson AI Lab \\
  \And
  Jonathan Ragan-Kelley \\
  MIT CSAIL
}

\begin{document}

\maketitle

\begin{comment}
    Mention the word "autoregressive"
\end{comment}
\begin{abstract}
    Key-value (KV) caching plays an essential role in accelerating decoding for transformer-based autoregressive large language models (LLMs). However, the amount of memory required to store the KV cache can become prohibitive at long sequence lengths and large batch sizes. Since the invention of the transformer, two of the most effective interventions discovered for reducing the size of the KV cache have been Multi-Query Attention (MQA) and its generalization, Grouped-Query Attention (GQA). MQA and GQA both modify the design of the attention block so that multiple query heads can share a single key/value head, reducing the number of distinct key/value heads by a large factor while only minimally degrading accuracy. In this paper, we show that it is possible to take Multi-Query Attention a step further by also sharing key and value heads between adjacent layers, yielding a new attention design we call Cross-Layer Attention (CLA). With CLA, we find that it is possible to reduce the size of the KV cache by another $2\times$ while maintaining nearly the same accuracy as unmodified MQA. In experiments training 1B- and 3B-parameter models from scratch, we demonstrate that CLA provides a Pareto improvement over the memory/accuracy tradeoffs which are possible with traditional MQA, enabling inference with longer sequence lengths and larger batch sizes than would otherwise be possible.
    % We open-source all of our code for replication. \footnote{\url{https://github.com/ibm-granite/dolomite-engine}}
\end{abstract}

\section{Introduction}

The memory footprint of the key-value (KV) cache can be a bottleneck when serving large language models (LLMs). Because the size of the KV cache scales proportionally with both sequence length and batch size, the memory overhead of KV cache storage can limit batch sizes when operating on long sequence lengths \citep{chowdhery2022palm}, and can require employing costly techniques like offloading when on-device memory is scarce \citep{flexgen}. It is also desirable to be able to persist KV caches over long periods of time in order to minimize redundant computations \citep{gao2024attentionstore,googleai2024context}. However, the size of the KV cache directly determines the cost of storing and retrieving such persistent caches. As new applications of LLMs emerge which demand ever-longer sequence lengths, the memory footprint of the KV cache is becoming an increasingly important consideration in the design of efficient transformer-based language models.

Existing work has proposed a variety of methods for decreasing the memory footprint of the KV cache, including storing KV activations in low precision \citep{hooper2024kvquant,zhang2024kvonebit}, evicting unimportant KV cache entries \citep{h2o,liu2023scissorhands}, and sharing keys and values across query heads in the attention mechanism \citep{shazeer2019mqa,ainslie2023gqa}.

% In this paper, we introduce \emph{Cross-Layer Attention}, a new architectural intervention for transformer models which reduces the size of the KV cache by sharing KV activations \emph{across layers}. Our contributions are as follows:
In this paper, we introduce a method for reducing the size of the KV cache along a dimension different than those explored in prior work: namely, reducing the number of unique \emph{layers} in the KV cache. Our contributions are as follows:
\begin{enumerate}
    \item We propose \emph{Cross-Layer Attention} (CLA), a modification to the transformer architecture which reduces the size of the KV cache by \emph{sharing KV activations across layers}.
    \item We conduct extensive pretraining experiments to characterize the effect of different CLA configurations on accuracy and memory usage across a range of architectural hyperparameters, learning rates and model sizes.
    \item We demonstrate that CLA enables accuracy/memory Pareto improvements relative to existing Multi-Query Attention (MQA) and Grouped-Query Attention (GQA) architectures.
    \item In particular, we demonstrate at the 1B- and 3B-parameter scales that combining CLA with MQA can achieve a $2\times$ reduction in KV cache size versus a plain MQA baseline, with minimal degradation in perplexity.
    \item We offer guidance on which CLA configurations perform best based on our experiments, finding that CLA should be used between pairs of consecutive layers, and that CLA appears to deliver the most robust benefits when used in conjunction with MQA.
\end{enumerate}

% In this paper, we investigate a new architectural intervention for decreasing the size of the KV cache which is orthogonal to the methods proposed by prior work: sharing KV activations \emph{across layers}.

% In this paper, we propose a method for reducing the size of the KV cache along an axis orthogonal to those explored in prior work: shrinking the KV cache in \emph{depth}. Specifically, we propose \emph{Cross-Layer Attention} (CLA), a modification to the transformer architecture which reduces the size of the KV cache by sharing KV activation \emph{across layers}. Our contributions are as follows:
% \begin{enumerate}
%     \item We propose Cross-Layer 
% \end{enumerate}

% In this paper, we propose \emph{Cross-Layer Attention}, a new architectural intervention for reducing the the size of the KV cache which works by sharing KV activations \emph{across layers}.

% https://ai.google.dev/gemini-api/docs/caching

\section{Cross-Layer Attention}

In this section we describe our Cross-Layer Attention (CLA) technique, and its relationship to the KV-sharing mechanisms employed by the existing Multi-Query and Grouped-Query attention architectures (MQA and GQA).

\begin{figure}
    \centering
    \includegraphics[scale=0.6]{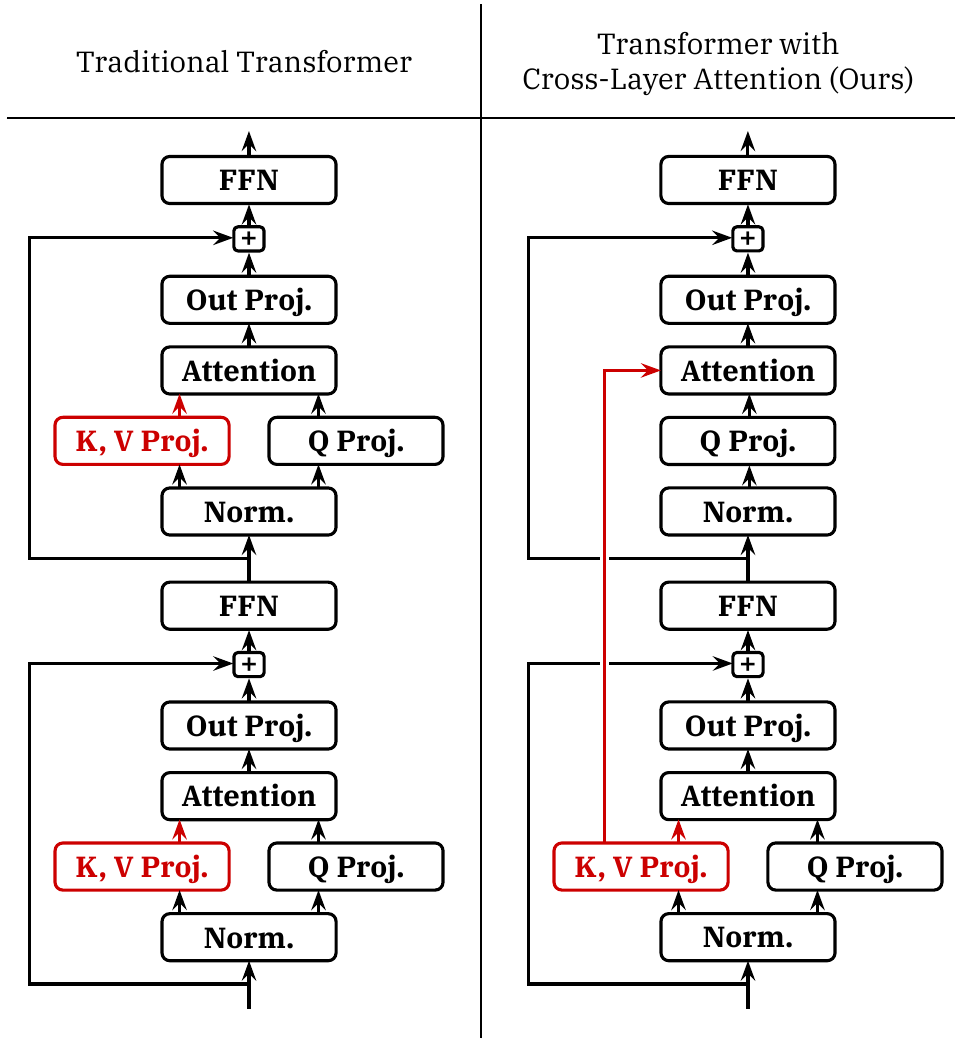}
    \caption{Schematic of two consecutive layers in a transformer using a traditional attention design (left) and in a transformer using Cross-Layer Attention (right).
    When using traditional attention, each layer computes its own separate $K$ and $V$ activations, which must be cached on a per-layer basis during autoregressive decoding.
    When using Cross-Layer Attention, some layers compute their own fresh $K$ and $V$ activations, while other layers reuse the $K$ and $V$ activations of earlier layers.
    }
    \label{fig:cla_block}
\end{figure}

\begin{figure}
    \centering
    \includegraphics[scale=0.6]{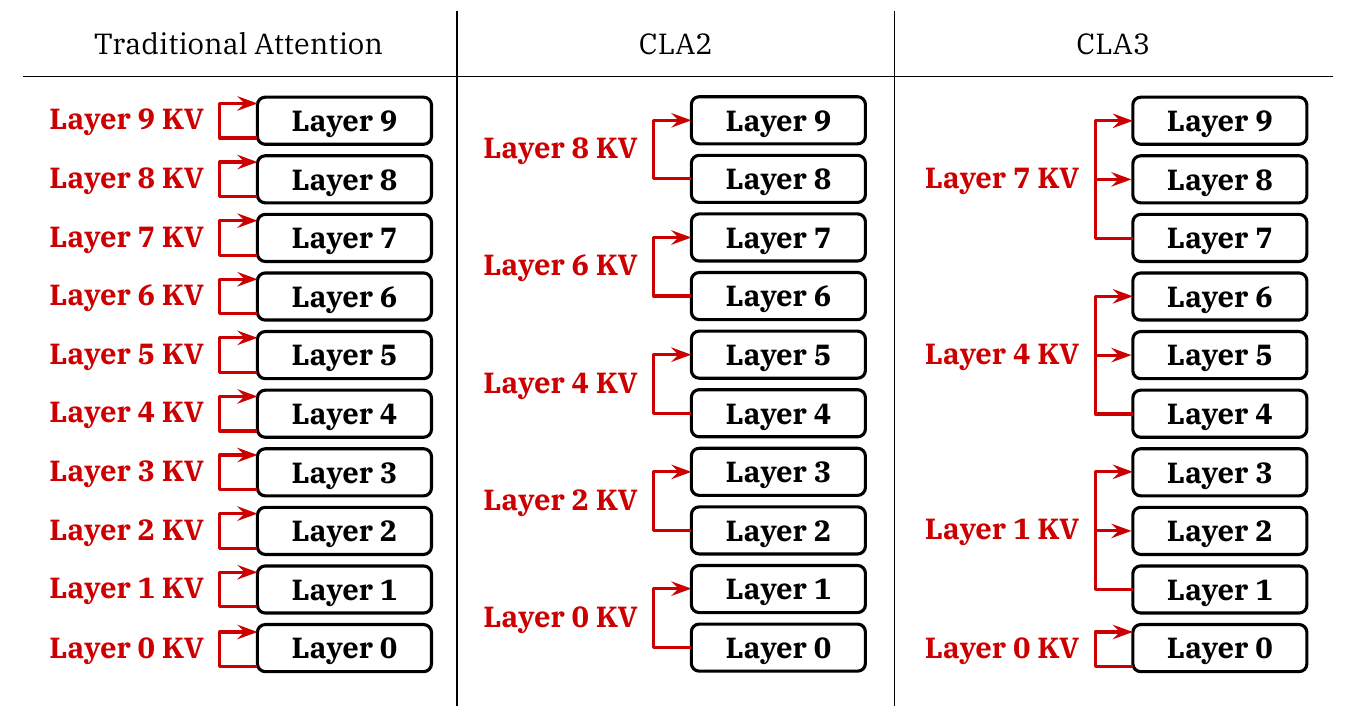}
    \caption{Schematic of KV cache structures under different attention configurations in a 10-layer transformer. Using traditional attention, each layer has its own KV cache. Using Cross-Layer Attention with a sharing factor of $2$ (CLA2), every group of $2$ consecutive layers shares a single KV cache. Using Cross-Layer Attention with a sharing factor of $3$ (CLA3), every group of $3$ consecutive layers shares a single KV cache. When the sharing factor does not evenly divide the number of layers, as in the CLA3 example, some KV caches must be shared over fewer layers than others; in this CLA3 configuration, we arbitrarily select the layer 0 KV cache to be used only in layer 0.}
    \label{fig:cla_layers}
\end{figure}

\subsection{Background: Multi-Query Attention and Grouped-Query Attention}

The original transformer architecture employed Multi-Head Attention (MHA) \citep{Vaswani+2017}, in which each query head attends over the keys and values produced by a distinct key/value head. In MHA, the KV activations of each key/value head must be stored separately in the KV cache, resulting in a storage overhead of $2 \cdot n_\text{query} \cdot d_\text{head}$ elements per token, where $n_\text{query}$ is the number of query heads and $d_\text{head}$ is the embedding dimension of each head.

To reduce the overhead associated with storing and accessing the KV cache during transformer decoding, \cite{shazeer2019mqa} proposed \emph{Multi-Query Attention} (MQA), which \citeauthor{ainslie2023gqa} later generalized to \emph{Grouped-Query Attention} (GQA). Grouped-Query Attention modifies the transformer architecture by organizing the query heads of each attention layer into groups, where each group of query heads shares a single key/value head. Because the size of the KV cache scales only with the number of distinct key/value heads, not the number of query heads, GQA reduces the storage overhead of the KV cache to $2 \cdot n_\text{group} \cdot d_\text{head}$, where $n_\text{group}$ denotes the number of groups for GQA and $n_\text{group} < n_\text{query}$. MQA can be seen as the special case of GQA in which $n_\text{group} = 1$.

\citeauthor{shazeer2019mqa} and \citeauthor{ainslie2023gqa} find that MQA and GQA enable significant reductions in KV cache size and decoding latency while incurring only a small degradation in accuracy compared to MHA architectures with the same head dimension. The family of attention architectures enabled by using MQA and GQA defines an \emph{accuracy/memory tradeoff space} in which model designers can choose how they want to balance the expressive power and KV cache overhead of their attention mechanism. MQA and GQA stake out different positions in this tradeoff space, and neither is necessarily preferable to the other for all use cases.

\subsection{Sharing KV Activations Across Layers}

Inspired by the success of MQA and GQA, which share key/value heads across query heads within a single layer, we propose also sharing key/value heads \emph{across layers}. We refer to such an attention architecture as \emph{Cross-Layer Attention} (CLA), and present a diagrammatic view of it in Figure \ref{fig:cla_block}. CLA computes key/value projections for only a subset of layers in the model; the attention blocks in layers without key/value projections reuse the KV activations of previous layers. Only the subset of layers with key/value projections contribute to the KV cache, allowing a reduction in memory footprint relative to traditional architectures which apply a separate key/value projection in each layer.

CLA is orthogonal to MQA/GQA/MHA, and can be combined with any of them. Moreover, in the same way that GQA allows varying $n_\text{group}$ to access a family of different attention configurations, CLA allows varying the number of layers which share the output of each KV projection, which we refer to as the \emph{sharing factor}. We refer to different configurations of CLA by their sharing factors, giving rise to CLA2, which shares each KV projection among a pair of adjacent layers, CLA3, which shares each KV projection among a group of $3$ layers, and so on. We present a visualization of different attention configurations possible with CLA in Figure \ref{fig:cla_layers}.

\subsection{Implications for System Design}

CLA is primarily an intervention to reduce the memory footprint of the KV cache, and only has minor effects on other resources consumed by the model during training and inference. Here, we summarize the effect of CLA on key metrics relevant from a systems engineering perspective, assuming all other architectural hyperparameters are held constant:
\begin{itemize}
    \item \textbf{KV Cache Memory}: CLA significantly reduces KV cache memory footprint, shrinking it by a factor equal to the sharing factor, or slightly less if the sharing factor does not evenly divide the number of layers.
    \item \textbf{Training Memory Footprint}: CLA reduces the memory footprint of intermediate KV activation tensors materialized during training, although for GQA and MQA models such KV tensors are typically small compared to the model's hidden states and MLP activations.
    \item \textbf{Model Parallelism}: CLA is fully compatible with standard tensor parallelism techniques \citep{shoeybi2020megatronlm} for sharding model weights across multiple accelerators. In the presence of pipeline parallelism \citep{10.5555/3454287.3454297}, either different layers which share a KV cache must be kept in the same pipeline stage, or else KV activations must be communicated between pipeline stages.
    % CLA doesn't change the transformer architecture significantly and thus is easily usable with Tensor parallelism \citep{shoeybi2020megatronlm} during both training and inference when the model might not fit on a single accelerator.
    \item \textbf{Parameters and FLOPs}: Because CLA reduces the total number of key/value projection blocks in the model, CLA slightly reduces the number of parameters in the model and the number of FLOPs required during a forward or backward pass.
    \item \textbf{Decoding Latency}: In the context of a full LLM serving stack, CLA can enable larger batch sizes and longer KV cache persistence times than would otherwise be possible, which have the potential to improve inference latency.
    \item \textbf{Core Attention Latency}: Unlike MQA and GQA, CLA has no direct effect on the memory bandwidth consumed by the attention mechanism in each decoding step, because even shared KV cache layers must be separately re-read from main memory in each attention layer. CLA therefore has no direct effect on the latency of the core attention computation during decoding.
\end{itemize}

\section{Pretraining Experiments}

\begin{figure}
    \centering
    \includegraphics[scale=1.0]{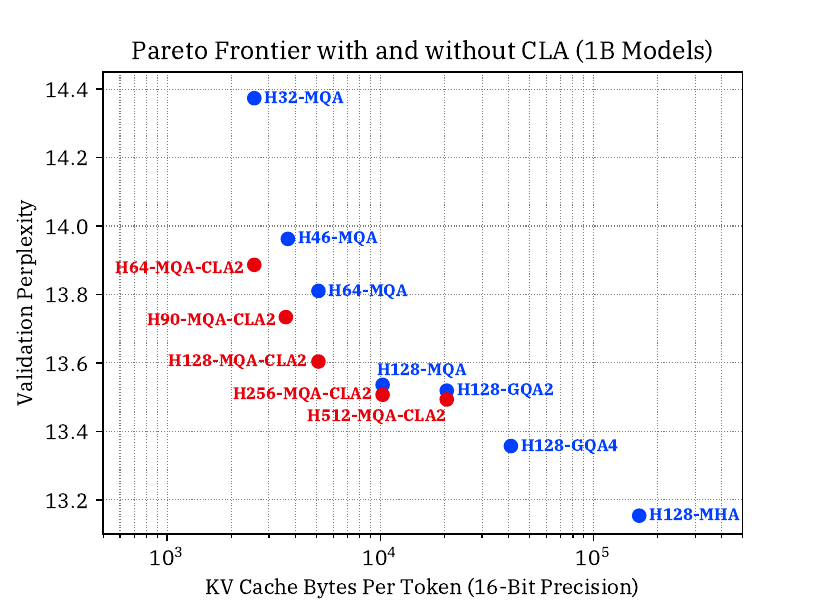}
    \caption{The accuracy/memory Pareto frontier discovered in our 1B-scale design space exploration, for models with CLA (red) and without CLA (blue). Lower is better on both axes.}
    \label{fig:1b_pareto}
\end{figure}

\begin{table}[]
    \centering
    
    \begin{tabular}{
        l
        S[table-format=3.0]
        S[table-format=2.0]
        S[table-format=2.0]
        S[table-format=2.0]
        S[table-format=5.0]
        S[table-format=2.2]
    }
        \toprule
        {\thead{\textbf{Model}}} &
        {\thead{${\boldsymbol{d}_{\textbf{head}}}$}} &
        {\thead{\textbf{Query}\\ \textbf{Heads}}} &
        {\thead{\textbf{KV}\\ \textbf{Heads}}} &
        {\thead{\textbf{KV}\\ \textbf{Layers}}} &
        {\thead{\textbf{KV Bytes Per}\\ \textbf{Token (16-Bit)}}} &
        {\thead{\textbf{Validation}\\ \textbf{Perplexity}}}
     \\
        \midrule
        \multicolumn{2}{l}{\textbf{Non-CLA Baselines}} \\
        \midrule
        H128-MHA & 128 & 16 & 16 & 20 & 163840 & 13.15 \\
        H128-GQA4 & 128 & 16 & 4 & 20 & 40960 & 13.36 \\
        H128-GQA2 & 128 & 16 & 2 & 20 & 20480 & 13.52 \\
        H128-MQA & 128 & 16 & 1 & 20 & 10240 & 13.54 \\
        H64-MQA & 64 & 32 & 1 & 20 & 5120 & 13.81 \\
        H46-MQA & 46 & 45 & 1 & 20 & 3680 & 13.96 \\
        H32-MQA & 32 & 64 & 1 & 20 & 2560 & 14.37 \\
        \midrule
        \multicolumn{2}{l}{\textbf{MQA + CLA2 Models}} \\
        \midrule
        H512-MQA-CLA2 & 512 & 4 & 1 & 10 & 20480 & 13.49 \\
        H256-MQA-CLA2 & 256 & 8 & 1 & 10 & 10240 & 13.51 \\
        H128-MQA-CLA2 & 128 & 16 & 1 & 10 & 5120 & 13.60 \\
        H90-MQA-CLA2 & 90 & 22 & 1 & 10 & 3600 & 13.73 \\
        H64-MQA-CLA2 & 64 & 32 & 1 & 10 & 2560 & 13.89 \\
        \midrule
        \multicolumn{2}{l}{\textbf{GQA + CLA2 Models}} \\
        \midrule
        H256-GQA4-CLA2 & 256 & 8 & 4 & 10 & 40960 & 13.38 \\
        H128-GQA4-CLA2 & 128 & 16 & 4 & 10 & 20480 & 13.48 \\
        H128-GQA2-CLA2 & 128 & 16 & 2 & 10 & 10240 & 13.59 \\
        \midrule
        \multicolumn{2}{l}{\textbf{MQA + CLA $\mathbf{> 2}$ Models}} \\
        \midrule
        H128-MQA-CLA3 & 128 & 16 & 1 & 7 & 3584 & 13.77 \\
        H128-MQA-CLA4 & 128 & 16 & 1 & 5 & 2560 & 13.95 \\
        \midrule
        \multicolumn{2}{l}{\textbf{MQA + CLA2, Non-Uniform Sharing}} \\
        \midrule
        H128-MQA-CLA2-KeepEnds & 128 & 16 & 1 & 11 & 5632 & 13.62 \\
        H128-MQA-CLA2-DenseFront & 128 & 16 & 1 & 11 & 5632 & 13.75 \\
        H128-MQA-CLA2-DenseBack & 128 & 16 & 1 & 11 & 5632 & 14.03 \\
    \bottomrule
    \end{tabular}
    
    \caption{Results of our 1B-scale design space exploration.}
    \label{tab:experiments_1b}
\end{table}

\begin{comment}
WB:
- To investigate the effect of cross-layer attention on transformer design, we trained a collection of transformer-based language models from scratch at the 1 billion and 3 billion parameter scales.
- Questions we wanted to answer:
    - What quality/memory tradeoffs is CLA able to achieve?
    - How do the tradeoffs enabled by CLA compare to the tradeoffs possible with existing MHA/GQA/MQA architectures?
    - When should you use CLA, and how should you combine it with other attention mechanisms?

- Question we wanted to answer:
    - What quality/memory tradeoffs does CLA enable?
    - How does CLA compare to plain GQA/MQA?
    - How does CLA interact with GQA/MQA and other hyperparameters?
    - Are the effects of CLA consistent across scales?
\end{comment}

To determine the effect of Cross-Layer Attention on language modeling accuracy, we trained a collection of transformer-based language models from scratch at the $1$ billion and $3$ billion parameter scales.
While running these experiments, we sought to answer the following questions:
\begin{enumerate}
    \item What accuracy/memory tradeoffs are possible using CLA?
    \item How does using CLA compare to using plain GQA or MQA?
    \item How does CLA interact with GQA and MQA?
    \item What CLA configurations perform best given a fixed memory budget?
    \item Are the effects of CLA consistent across scales?
\end{enumerate}
We found that CLA enables favorable accuracy/memory tradeoffs compared to what is possible using plain GQA or MQA.
Moreover, we found that in our experimental regime, a sharing factor of $2$ is more effective than other sharing factors, and that CLA is consistently effective when combined with MQA when trying to decrease KV cache storage.
We also found preliminary evidence to suggest that CLA models benefit from training with higher learning rates than comparable non-CLA models.
Finally, we found that CLA confers benefits at both 1B- and 3B-parameter scales.
In the rest of this section, we present our experimental setup and results in more detail.

\subsection{Common Experimental Parameters}

In all our experiments, we train our models from scratch on data from the SlimPajama \citep{cerebras2023slimpajama} dataset, tokenized with the GPT-NeoX tokenizer \citep{black2022gptneox20b} which uses Byte-Pair Encoding (BPE) \citep{wang2019neural}.
We adopt a Llama-like \citep{touvron2023llama} architecture with pre-normalization, SwiGLU activations \citep{shazeer2020glu, ramachandran2017searching}, and rotary position embeddings \citep{su2023roformer}.
We do not use dropout for any of our models. Our models include learnable elementwise affine parameters for layer-norm, and our CLA models use separately-learnable affine layer-norm parameters for the KV projection blocks and Q projection blocks in attention. Unless otherwise stated, we always set the number of query heads $n_\text{query}$ such that $n_\text{query} \cdot d_\text{head}$ is equal to the hidden size $d_\text{model}$.

We train all models using the AdamW optimizer \citep{loshchilov2018decoupled} with gradient clipping, using $\beta_1 = 0.9$, $\beta_2 = 0.95$, a weight decay factor of $0.1$, and a clipping norm of $1.0$.
We use a linear learning rate warmup for the first $5\%$ of training examples and a cosine learning rate schedule \cite{loshchilov2017sgdr} decaying to $10\%$ of the peak learning rate over the remainder of training.
We set the sequence length to $2048$ tokens and the batch size to $2048$ sequences, for a total of $\approx 4M$ tokens per training step.
All our experiments initialize the weights of linear layers from a normal distribution with mean zero and standard deviation $0.01275$.

We perform all experiments on NVIDIA H100 GPUs using PyTorch \citep{paszke2019pytorch, Ansel_PyTorch_2_Faster_2024}. We use mixed precision training \citep{micikevicius2018mixed} in BF16 \citep{kalamkar2019study} with gradient all-reduce and gradient accumulation in FP32 for training stability.

\subsection{Experiments at 1B-Parameter Scale}

\begin{table}[]
    \centering
    \begin{tabular}{@{}l|S[table-format=4.0]S[table-format=4.0]S[table-format=2.0]S[table-format=4.0]S[table-format=3.0e2]@{}}
    \toprule
    \textbf{Model Family} & \textbf{Hidden Size} & \textbf{FFN Size} & \textbf{Layers} & \textbf{Sequence Length} & \textbf{Training Tokens} \\
    \midrule
    1B Models & 2048 & 5472 & 20 & 2048 & 30e9 \\
    3B Models & 3072 & 8192 & 32 & 2048 & 100e9 \\
    \bottomrule
    \end{tabular}
    \title{Hello}
    \caption{Architectural and training hyperparameters shared across our pretraining experiments.}
    \label{tab:arch_hyperparams}
\end{table}

% We ran our most comprehensive set of experiments at the 1B-parameter scale. 
We trained all our 1B-scale models on 30 billion tokens using a consistent data order, and, other than varying the attention mechanism, used the same architectural hyperparameters across all 1B-scale models.
This means that all our 1B models were all trained using approximately the same number of FLOPs and approximately the same number of GPU-hours, with CLA models requiring slightly fewer FLOPs to train than their non-CLA counterparts due to the reduced number of key/value projections.
% check: is this stated twice
The common hyperparameters shared across our 1B-scale experiments can be found in Table \ref{tab:arch_hyperparams}.

We ran two main sets of experiments at the 1B-parameter scale. First, we trained a diverse set of CLA and non-CLA models to characterize the range of accuracy/memory tradeoffs achievable with and without CLA, and to determine which CLA configurations are most effective; we refer to these as our \emph{design space exploration} experiments, and describe them in more detail in Section \ref{sec:1b_design_space_exploration}.
Second, we conducted a learning rate sweep on a subset of models from our design space exploration to verify that our results continue to hold even against a strong non-CLA baseline with a well-tuned learning rate.
We describe these learning rate tuning experiments in Appendix \ref{sec:lr-sweep}.

\subsubsection{Design Space Exploration}\label{sec:1b_design_space_exploration}

The primary goal of our 1B-parameter-scale design space exploration was to characterize the Pareto frontier of accuracy/memory tradeoffs achievable with and without CLA, and to determine which CLA configurations achieve the best accuracy on a fixed KV cache memory budget.
We train all models in our design space exploration using a learning rate of $\text{LR} = 3\times10^{-4}$, which we determined to be conservative; we explore the effect of the learning rate on accuracy in more detail in Section \ref{sec:1b_lr_sweep}.

For our design space exploration, we first trained a collection of seven non-CLA baseline models along the MHA-GQA-MQA spectrum, exhibiting a range of KV cache memory requirements spanning two orders of magnitude.
Our baseline model with the largest KV cache memory footprint is an MHA model with a head embedding dimension of $d_\text{head} = 128$ ($163840$ bytes per token at 16-bit precision), and our baseline with the smallest footprint is an MQA model with head dimension $d_\text{head} = 32$ ($2560$ bytes per token).

We quantify the accuracy of models in our design space exploration using perplexity on a held-out validation set of $\approx 4M$ tokens drawn from our SlimPajama corpus.
A summary of results for the models in our design space exploration, including our baseline models, can be found in Table \ref{tab:experiments_1b}. We adopt the naming scheme ``H$\langle d_\text{head} \rangle$-$\langle \text{attention mechanism} \rangle$'' for all models in our experiments, so that, for example, a model employing MQA with a head dimension of $d_\text{head} = 64$ would be named ``H64-MQA.''
For our baseline models, we observe that validation perplexity increases monotonically as we reduce the memory capacity of the KV cache, ranging from a perplexity of $13.15$ for our H128-MHA baseline to $14.37$ for our H32-MQA baseline.

In the rest of this section, we present results for the CLA models we trained during our design space exploration.

% Paragraph is clunky, use more sentences
\paragraph{Best Performance: MQA + CLA2.} We trained a family of five models combining MQA with CLA2.
%(across layers) using a sharing factor of $2$ for CLA, yielding an attention mechanism we refer to as ``MQA-CLA2.''
% Because our 1B-parameter models have $20$ transformer layers, and CLA2 computes keys and values in only half the layers in the model, our MQA-CLA2 models each have $10$ layers of KV cache, and each MQA-CLA2 model requires only half the KV cache memory of a corresponding MQA model with the same head dimension $d_\text{head}$.
We varied the head dimension for our MQA-CLA2 models from $d_\text{head} = 512$ down to $d_\text{head} = 64$, allowing us to compare to a range of non-CLA baseline models with varying KV cache capacities.

\begin{comment}
We find that our MQA-CLA2 models are able to achieve better perplexities than baseline models requiring the same amount of KV cache memory, advancing the accuracy/memory Pareto frontier.
We present a plot of the accuracy/memory Pareto frontier with and without CLA in Figure \ref{fig:1b_pareto}.
The CLA-enabled models H128-MQA-CLA2, H90-MQA-CLA2, and H64-MQA-CLA2 match the KV cache footprints of the baseline models H64-MQA, H46-MQA, and H32-MQA, respectively, while achieving absolute perplexity improvements of $0.21$, $0.23$, and $0.48$ points.
Additionally, by expanding the head dimension to values greater than our standard head dimension of $d_\text{head} = 128$, we obtain the CLA-enabled models H256-MQA-CLA2 and H512-MQA-CLA2, which match the KV cache footprints of the baseline models H128-MQA and H128-GQA2, respectively, while achieving a validation perplexity improvement of $0.03$ in both cases.\todo[size=\small]{Is this statistically significant? How should we even quantify statistical significance here?}
\end{comment}

We found that our MQA-CLA2 models are able to achieve better perplexities than baseline models requiring the same amount of KV cache memory, advancing the accuracy/memory Pareto frontier.
We present a plot of the accuracy/memory Pareto frontier with and without CLA in Figure \ref{fig:1b_pareto}. Our MQA-CLA2 models with head dimensions $d_\text{head} \in \{64, 90, 128\}$ are able to match the KV cache memory footprint of baseline MQA models with head dimensions $d_\text{head} \in \{32, 46, 64\}$ while achieving substantial perplexity improvements in the range of $0.21$--$0.48$ points. Additionally, our MQA-CLA2 models with large head sizes of $d_\text{head} \in \{256, 512\}$ are able to match the KV cache footprint of our MQA and GQA2 baselines with $d_\text{head} = 128$ while achieving a small perplexity improvement of $0.03$ points.
% We found that our MQA-CLA2 models are able to match the KV cache memory footprint of baseline MQA models 
% with small head dimensions 
% ($d_\text{head} \in \{32, 46, 64\}$) while achieving substantial perplexity improvements in the range of $0.21$--$0.48$ points. Additionally, we found that MQA-CLA2 models with large head sizes of $d_\text{head} \in \{256, 512\}$ are able to match the KV cache footprint of our $d_\text{head} = 128$ MQA and GQA2 baselines while achieving a small perplexity improvement of $0.03$ points.

We found that our MQA-CLA2 models achieved the best accuracy/memory tradeoffs among all CLA configurations we tested in our design space exploration. In the rest of this section, we briefly describe the ablations we conducted to explore alternate CLA configurations.

\paragraph{Ablation: GQA + CLA2.}

We trained three models to explore combining GQA with CLA2.
We chose GQA4-CLA2 with $d_\text{head} = 128$ as our starting point, as GQA4 represents an attention configuration intermediate between our MQA and MHA baselines. We then explored expanding the head dimension of our GQA4-CLA2 model to $d_\text{head} = 256$, as well as reducing the GQA factor to GQA2. We found that only the GQA2-CLA2 configuration was able to achieve a perplexity better than the corresponding baseline model with the same KV cache footprint, and that this perplexity was the same (within $0.01$ points) as our MQA-CLA2 model with the same footprint.

% We trained three models combining GQA (as opposed to MQA) with CLA2: H256-GQA4-CLA2, H128-GQA4-CLA2, and H128-GQA2-CLA2, which match the KV cache footprints of the baseline models H128-GQA4, H128-GQA2, and H128-MQA, respectively. We found that the models H256-GQA4-CLA2 and H128-GQA2-CLA2 achieved worse perplexities than the corresponding baseline models, and that H128-GQA4-CLA2 achieved nearly the same perplexity ($13.48$ vs $13.49$) as the MQA-CLA2 model H512-MQA-CLA2, which has the same KV cache footprint.

% We trained three models combining GQA with CLA2. 
% We found that two of these models were worse than their corresponding baselines, while one performed similarly to MQA-CLA2, as ween in CITE TABLE.

% \paragraph{Ablation: MQA + CLA with Sharing Factor $\mathbf{> 2}$.} We trained two models combining MQA with CLA using a sharing factor $> 2$ for CLA: H128-MQA-CLA3, which uses a sharing factor of $3$, and H128-MQA-CLA4, which uses a sharing factor of $4$. For H128-MQA-CLA3, we share the first-layer KV cache over only the first $2$ layers of the model rather than the first $3$, because it is not possible to evenly divide a $20$-layer model into groups of layers of size $3$. The models H128-MQA-CLA3 and H128-MQA-CLA4 have approximately the same KV cache footprints as the baselines H46-MQA and H32-MQA, respectively, and as the CLA2 models H90-MQA-CLA2 and H64-MQA-CLA2, respectively. We found that H128-MQA-CLA3 and H128-MQA-CLA4 both outperform their corresponding baselines, but underperform their corresponding CLA2 models.

\paragraph{Ablation: MQA + CLA with Sharing Factor $\mathbf{> 2}$.} To explore the effect of using CLA sharing factors $> 2$, we trained MQA-CLA3 and MQA-CLA4 models with head dimension $d_\text{head} = 128$. We found that these CLA3 and CLA4 models achieved a Pareto improvement over our plain MQA baselines, matching the KV cache footprint of our baseline MQA models with head dimensions of $d_\text{head} \in \{32, 46\}$ while achieving better perplexities. However, we found that they achieved worse perplexities than our MQA-CLA2 models at the same KV cache footprint.

\paragraph{Ablation: MQA + CLA2 with Non-Uniform Sharing Patterns.} Finally, we explored using different patterns of KV activation sharing in our MQA-CLA2 models.

On the hypothesis that the first and last layers in the model might benefit from special treatment, we trained a model ``H128-MQA-CLA2-KeepEnds'' which does not share the layer $0$ KV cache with any other layers, and instead groups layer $1$ with layer $2$, groups layer $3$ with layer $4$, and so on. This also has the effect of giving the final layer its own KV cache separate from all other layers.

We also explored imbalanced configurations with all the KV-cache-producing layers concentrated at either the beginning or end of the model. We trained a model ``H128-MQA-CLA2-DenseFront'' consisting of $10$ non-CLA layers, followed by $9$ CLA layers all using the KV activations of layer $9$, and a final layer with its own KV cache. Similarly, we trained a model ``H128-MQA-CLA2-DenseBack'' consisting of $2$ non-CLA layers, followed by a run of $10$ CLA layers all using the KV activations of layer $1$, and finally $9$ non-CLA layers.

We found that all of these alternative CLA sharing patterns achieve worse perplexities than the corresponding MQA-CLA2 model with a uniform sharing pattern, while also requiring slightly more KV cache memory.

\begin{comment}
WB:
- Things we might want to say:
    - 128 is in some way the "natural" head dimension
    - Compared to shrinking below 128, we do great
    - Compared to 128, we do pretty well in not incurring a big accuracy hit from shrinking 2x
    - We obtain very modest memory-matched accuracy improvements over H128-MQA and H128-GQA2
\end{comment}

% \textbf{Marginal Result: GQA+CLA2 vs GQA.}

% \textbf{Negative Result: CLA with Sharing Factor $> 2$.}

% \textbf{Negative Result: CLA2 with Non-Uniform Sharing Patterns.}

\subsubsection{Robustness to Learning Rate Tuning}\label{sec:1b_lr_sweep}

The relative performance of different model architectures can change depending on the learning rates at which they are evaluated. To account for the effects of the learning rate on our results, we conducted learning rate tuning experiments on three models of interest from our initial 1B-scale design space exploration. These learning rate tuning experiments help us verify that CLA continues to provide benefits even when compared to baselines trained at their optimal learning rates.

We chose to tune the learning rate for the baseline models H128-MQA and H64-MQA, as well as the CLA model H128-MQA-CLA2. In our initial design space exploration, our results for these models indicated that CLA makes it possible to shrink the KV cache footprint of an MQA model with $d_\text{head} = 128$ by a factor of $2\times$ while incurring only a small ($0.06$ point) degradation in perplexity, or to create a model with the same KV cache footprint as an MQA model with $d_\text{head} = 64$ while enjoying a substantial ($0.21$ point) improvement in perplexity. We wanted to verify that this qualitative pattern continues to hold when all models are trained with well-tuned learning rates.

\paragraph{Learning Rate Tuning Strategy.} For each of our three model configurations, we swept the learning rate upwards from an initial value of $3 \times 10^{-4}$ in multiplicative increments of $1.5\times$. We ended our sweep for each model at the point where validation perplexity stopped improving. We treat the learning rate which achieved the lowest validation perplexity for each model as an approximation of that model's optimal learning rate.

\paragraph{Results.} We found an optimal learning rate of $\text{LR} = 1.5 \times 10^{-3}$ for our H128-MQA baseline, and a higher optimal learning rate of $\text{LR} = 2.25 \times 10^{-3}$ for both our H64-MQA baseline and our H128-MQA-CLA2 model.

The validation perplexity results from our 1B-scale learning rate tuning experiments can be found in Table \ref{tab:1b_lrtune_perplexity}. When comparing all three models at their best learning rates, we found that the qualitative result from our design space exploration continues to hold: our CLA2 model incurs only a small ($0.04$ point) validation perplexity degradation relative to our $d_\text{head} = 128$ baseline while enjoying a $2\times$ smaller KV cache footprint, and achieves a substantial ($0.31$ point) validation perplexity improvement compared to our $d_\text{head} = 64$ baseline while using the same amount of KV cache memory.

To further validate our results, we also evaluate our three learning-rate-tuned 1B-scale models under EleutherAI's LM Eval Harness \citep{lm-eval-harness} on Wikitext \citep{merity2016pointer} perplexity  and seven standard downstream benchmarks. We report the results of these evaluation in tables \ref{tab:1b_lrtune_perplexity} and \ref{tab:1b_lrtune_downstream}. On Wikitext perplexity, we observe a similar pattern as with validation perplexity, with our tuned CLA2 model achieving approximately the same ($0.01$ points better) perplexity as our $d_\text{head} = 128$ baseline, and substantially ($0.71$ points) better perplexity than our $d_\text{head} = 64$ baseline. On the downstream evaluations, we found that none of our three models model consistently wins or loses across different benchmarks, and that all three models are consistently within $1$--$5$ percentage points of each other.

\begin{table}[h]
    \centering
    \begin{tabular}{lcccc}
        \hline
        \thead{\textbf{Model}} & \thead{\textbf{KV Bytes Per} \\ \textbf{Token (16-bit)}} & \thead{\textbf{Best LR}} & \thead{\textbf{Validation} \\ \textbf{Perplexity}} & \thead{\textbf{Wikitext} \\ \textbf{Perplexity}} \\
        \hline
        H128-MQA & 10240 & $1.5\phantom{0} \times 10^{-3}$ & $\mathbf{12.39}$ & $19.30$ \\
        H128-MQA-CLA2 & \phantom{0}5120 & $2.25 \times 10^{-3}$ & $12.43$ & $\mathbf{19.29}$ \\
        H64-MQA & \phantom{0}5120 & $2.25 \times 10^{-3}$ & $12.74$ & $20.00$ \\
        \hline
    \end{tabular}
    \caption{Perplexity results from learning rate tuning experiments at 1B-parameter scale.}
    \label{tab:1b_lrtune_perplexity}
\end{table}

\begin{table}[h]
    \centering
    \begin{tabular}{lccccccc}
        \hline
        \textbf{Model (Best LR)} & \textbf{Hellaswag} & \textbf{PIQA} & \textbf{WG} & \textbf{SciQ} & \textbf{OBQA} & \textbf{BoolQ} & \textbf{ARC-E} \\ \hline
        H128-MQA & \textbf{36.24} & 69.15 & \textbf{52.96} & \textbf{82.9} & 19.0 & \textbf{57.40} & \textbf{55.43} \\
        H128-MQA-CLA2 & 36.01 & 69.15 & 51.93 & 82.6 & \textbf{21.4} & 53.21 & 53.87 \\
        H64-MQA & 35.22 & \textbf{69.21} & 50.75 & 78.5 & 19.4 & 55.81 & 51.68 \\
        \hline
    \end{tabular}
    \caption{Downstream benchmark results for 1B-scale models with tuned learning rates. The columns ``WG'' and ``OBQA'' denote ``WinoGrande'' and ``OpenBookQA'', respectively.}
    \label{tab:1b_lrtune_downstream}
\end{table}

\subsection{Experiments at 3B-Parameter Scale}

To determine how CLA performs when applied to larger models, we trained a collection of models at the 3B-parameter scale both with and without CLA. We trained each of our 3B-scale model from scratch on 100B tokens from our SlimPajama corpus. The common architectural hyperparameters for our 3B-scale models can be found in Table \ref{tab:arch_hyperparams}.

\paragraph{Experiments at Head Dimension $\boldsymbol{d}_\textbf{head} = \mathbf{128}$.} We initially ran experiments to compare three 3B-scale models analogous to the models we selected for our learning rate tuning experiments at the 1B-parameter scale. Specifically, we compared a model using MQA-CLA2 and $d_\text{head} = 128$ to an MQA model with the same head dimension (and hence $2\times$ the KV cache footprint), and to an MQA model with a head dimension of $d_\text{head} = 64$ (and hence the same KV cache footprint). Based on our 1B-scale experiments, we expected that our MQA-CLA2 and MQA models with $d_\text{head} = 128$ would achieve similar perplexities to each other, and that both would outperform the $d_\text{head} = 64$ model.

We tuned the learning rates for these models according to the same learning rate tuning protocol we used at the 1B-parameter scale. After tuning the learning rates for each model, we observed a result different than we had expected: at 3B scale, our MQA-CLA2 model achieves substantially better perplexities than both our $d_\text{head} = 128$ and $d_\text{head} = 64$ MQA baselines. Moreover, our $d_\text{head} = 64$ MQA baseline model achieves better perplexities than our tuned $d_\text{head} = 128$ MQA baseline, despite having only $\nicefrac{1}{2}$ as much KV cache capacity. We report the optimal learning rates and perplexities for these three models in Table \ref{tab:3b_128_lrtune_perplexity}.

As with our 1B-scale learning rate tuning experiments, we evaluate these models on downstream benchmarks. We report the results of these evaluations in Table \ref{tab:3b_128_lrtune_downstream}. As with our 1B-scale experiments, we do not find that any model consistently wins or loses in these downstream evaluations.

\begin{table}[h]
    \centering
    \begin{tabular}{lcccc}
        \hline
        \thead{\textbf{Model}} & \thead{\textbf{KV Bytes Per} \\ \textbf{Token (16-bit)}} & \thead{\textbf{Best LR}} & \thead{\textbf{Validation} \\ \textbf{Perplexity}} & \thead{\textbf{Wikitext} \\ \textbf{Perplexity}}  \\
        \hline
        H128-MQA & 16384 & $6.75 \times 10^{-4}$ & $9.52$ & $13.63$ \\
        H128-MQA-CLA2 & \phantom{0}8192 & $2.25 \times 10^{-3}$ & $\mathbf{9.34}$ & $\textbf{13.25}$ \\
        H64-MQA & \phantom{0}8192 & $1.00 \times 10^{-3}$ & $9.48$ & $13.49$ \\
        \hline
    \end{tabular}
    \caption{Optimal learning rate and perplexity results for our first set of 3B-scale experiments.}
    \label{tab:3b_128_lrtune_perplexity}
\end{table}

\begin{table}[h]
    \centering
    \begin{tabular}{lccccccc}
        \hline
        \textbf{Model (Best LR)} & \textbf{Hellaswag} & \textbf{PIQA} & \textbf{WG} & \textbf{SciQ} & \textbf{OBQA} & \textbf{BoolQ} & \textbf{ARC-E} \\ \hline
        H128-MQA      & 45.73 & 73.07 & 60.46 & 88.1 & 25.4 & 59.30 & \textbf{64.90} \\
        H128-MQA-CLA2 & \textbf{47.12} & \textbf{74.32} & \textbf{60.69} & \textbf{89.2} & 25.2 & 58.62 & 64.73 \\
        H64-MQA       & 46.42 & 74.05 & 57.85 & 88.1 & \textbf{25.6} & \textbf{59.88} & 65.57 \\
        \hline
    \end{tabular}
    \caption{Downstream evaluation results for our first set of 3B-scale experiments.}
    \label{tab:3b_128_lrtune_downstream}
\end{table}

\paragraph{Experiments at Head Dimension $\boldsymbol{d}_\textbf{head} = \mathbf{64}$.} Because in our initial 3B-scale experiments we found that our $d_\text{head} = 64$ MQA model represents a stronger baseline than our $d_\text{head} = 128$ MQA model, we ran a second set of experiments with adjusted head sizes. Specifically, we chose to compare an MQA-CLA2 model with $d_\text{head} = 64$ to a plain MQA model with $d_\text{head} = 64$, and to a plain MQA model with $d_\text{head} = 32$.

Due to logistical constraints we trained all models in this second set of 3B-scale experiments on a different training cluster using a different training software stack and data order. This included retraining a new version of our H64-MQA-CLA2 baseline in order to control for differences in the new training environment.

We trained all models in this second set of experiments with a learning rate of $\text{LR} = 10^{-3}$, which we had found to be the best learning rate for our $d_\text{head} = 64$ MQA baseline model in our first set of 3B-scale experiments. For our $d_\text{head} = 64$ MQA-CLA2 model and our $d_\text{head} = 32$ MQA baseline model, we also experimented with learning rates of $\text{LR} \in \{6.75 \times 10^{-4}, 1.5 \times 10^{-3}\}$, but found these achieved worse perplexities than our initial value of $\text{LR} = 10^{-3}$.

We report perplexity results for this second set of experiments in Table \ref{tab:3b_64_lrtune_perplexity}, and results for downstream benchmarks in Table \ref{tab:3b_64_lrtune_downstream}. In the Wikitext perplexity results for this set of experiments, we find agreement with the pattern observed at the 1B scale. Our MQA-CLA2 model with $d_\text{head} = 64$ incurs only a small ($0.05$ point) degradation in perplexity compared to our $d_\text{head} = 64$ baseline while enjoying a $2 \times$ smaller KV cache footprint, and achieves a substantial ($0.35$ point) improvement in perplexity compared to our $d_\text{head} = 32$ baseline while using the same amount of KV cache memory.

We also evaluate these three models on downstream benchmarks, and report the results in Table \ref{tab:3b_64_lrtune_downstream}. As with our downstream benchmark evaluations for our other experiments, we find that all models perform similarly.

\begin{table}[h]
    \centering
    \begin{tabular}{lccc}
        \hline
        \thead{\textbf{Model}} & \thead{\textbf{KV Bytes Per} \\ \textbf{Token (16-bit)}} & \thead{\textbf{Best LR}} & \thead{\textbf{Wikitext Perplexity}} \\
        \hline
        H64-MQA & 8192 & $1.0 \times 10^{-3}$ & $\textbf{12.94}$ \\
        H64-MQA-CLA2 & 4096 & $1.0 \times 10^{-3}$ & $12.99$ \\
        H32-MQA & 4096 & $1.0 \times 10^{-3}$ & $13.34$ \\
        \hline
    \end{tabular}
    \caption{Optimal learning rate and perplexity results for our second set of 3B-scale experiments.}
    \label{tab:3b_64_lrtune_perplexity}
\end{table}

\begin{table}[h]
    \centering
    \begin{tabular}{lccccccc}
        \hline
        \textbf{Model (Best LR)} & \textbf{Hellaswag} & \textbf{PIQA} & \textbf{WG} & \textbf{SciQ} & \textbf{OBQA} & \textbf{BoolQ} & \textbf{ARC-E} \\ \hline
        H64-MQA      & \textbf{47.34} & \textbf{74.54} & \textbf{60.46} & \textbf{88.9} & 24.2 & 57.25 & \textbf{66.92} \\
        H64-MQA-CLA2 & 47.32 & \textbf{74.54} & 57.46 & 87.9 & 25.2 & 61.62 & 65.53 \\ 
        H32-MQA      & 46.05 & 73.83 & 60.06 & 88.6 & \textbf{25.6} & \textbf{61.87} & 65.24 \\
        \hline
    \end{tabular}
    \caption{Downstream evaluation results for our second set of 3B-scale experiments.}
    \label{tab:3b_64_lrtune_downstream}
\end{table}

% Similarly to the three models we selected for our learning rate tuning experiments at the 1B-parameter scale, at the 3B scale we initially sought to compare an MQA-CLA2 model with $$

\begin{comment}
Outline:
- Two groups of experiments:
    - 128/128/64
    - 64/64/32
        - Found results here more similar to 1B scale
\end{comment}

\section{Discussion \& Future Work}

\begin{comment}
Possible topics:
- Clear takeaway: CLA helps at all scales we tested
- When does CLA help?
    - Helps in conjunction with MQA but not GQA
    - Helps most with CLA2, higher factors counterproductive
    - Helps most with "standard" sizes
    - head dims
- Implications:
    - If you have a normally-scaled MQA model, you can probably get a free 2x KV cache size win with no significant hit to accuracy by plugging CLA2 into it
    - If you're not using MQA yet, you should do MQA before adding CLA2
    - If you're already using MQA and CLA2 and want to shrink KV cache size further, you should shrink the head size

Simple sentences:
- 1: We should scale up, put this in a real inference system under load, and show some end-to-end cost reductions
    - 1.1: Combine with interesting features where KV cache storage or networking costs are relevant
    - 1.2: (Related) we should scale up to long context
    - 1.3: (Related) we should try combining with other efficient attention techniques that reduce time costs of attention.
\end{comment}

\paragraph{Takeaways and Implications.} 
% We find that MQA-CLA2 consistently achieves the lowest (or near the lowest) validation perplexity for a given storage limitation and model size.
% We arrive at this result by finding that using greater degrees of CLA than two hurts more than using greater degrees of head parallelism(ex. 32 head H64-MQA-CLA2 is better 16 head H128-MQA-CLA4). All model with some degree of CLA outperform models with just greater head parallelism, however.
% From this study, both GQA4 and MQA both seem to perform well with CLA2, achieving less than 1\% degradation in perplexity when using the method across the board. However, MQA combined with CLA achieves the largest gain relative to the existing pareto frontier.
% As a result, we posit that most practitioners interested in the memory storage benefits of CLA should use it for a 2x decrease in memory storage on top of MQA.

In the regimes where we tested it, we find that MQA-CLA2 consistently achieves the lowest validation perplexity (within $0.01$ points) for a given KV cache memory budget and model size. In our ablations, we find that using sharing factors greater than $2$ (CLA3 and above) achieves slightly worse accuracy/memory tradeoffs than using CLA2 and varying the head dimension, although still Pareto-dominates the tradeoffs possible with plain MQA alone.

At both 1B and 3B scale, we find that for MQA models with typical head sizes of $64$ and $128$, applying CLA2 yields a $2\times$ KV cache reduction while incurring at worst a very modest (less than $1\%$ change) degradation in perplexity, and in some cases improving perplexity. We recommend this recipe to practitioners as a conservative change to existing MQA architectures which delivers substantial memory overhead reductions with relatively little risk.

\paragraph{Future Work.}
One natural question that rises from any memory efficient LLM alternative is its efficiency improvement when serving through longer sequences and greater batching.
We leave end-to-end inference efficiency evaluations of large, long-context models employing CLA as an interesting problem for future work.
We suspect that the types of LLMs that will stand to gain the most are the ones which have extremely long sequences, such as models that have long term memory or use methods like Landmark Attention \citep{landmark} which render attention over long contexts more feasible.

\section{Related Work}

Transformer memory efficiency can refer to many potential objectives. 
It can refer to decreasing memory storage or bandwidth requirements, it can be targeted at training or inference, and finally it apply either within a single pass of the model or between passes.
While this work targets decreasing the size of the inference KV cache that persists between passes, notable works such as 
Flash Attention \citep{dao2022flashattention, dao2023flashattention2} have decreased the memory bandwidth necessary for a single pass,
and works like FlexGen \citep{flexgen} achieved low memory storage during a single forward pass by partial offloading  to disk.
Here we discuss related work that improves the memory efficiency of attention, specifically the memory storage of the KV cache.

\subsection{Decreasing KV cache size Post training}
Much work has focused on decreasing the size of the KV cache for models that have already been trained.

\paragraph{KV cache compression}
As many works have tried to compress LLMs through pruning, quantization, and sparsity, 
(see \citet{zhu2023survey} for a survey)
a subset directly focus on the problem of KV cache compression.
For quantization,
KVQuant \citep{hooper2024kvquant}
and
Coupled Quantization \citep{zhang2024kvonebit}
perform targeted transformations of the keys and values along with non uniform encodings to compress the KV cache to one to two bits.
% cite other AQ methods?
Sparsifying the KV cache done by works such as 
H2O \citep{h2o}
Scissorhands
\citep{liu2023scissorhands}
and FastGen \citep{ge2024fastgen}
only store a subset of the KV cache during generation.
They do so by storing only tokens that are near to the generating token or important across the sequence, with the heuristic for importance varying between papers.
Finally, Cachegen \citep{liu2024cachegen} directly compresses the KV cache with a tensor encoder.

\subsection{Architectural Changes that decrease KV cache size} 

Most relevant to our work are methods that change the architecture of the model in order to decrease the size of the KV cache. 
These methods can roughly grouped into three categories:
those that try to reduce the number of tokens attended to,
those that replace softmax attention with another operation that requires less memory storage,
and those that, like our work, decrease the unique KV values compared to each query.
\paragraph{Decreasing effective sequence length}
Models that decrease the effective sequence length of the model have existed almost as transformers themselves.
Two notable early works are
Transformer XL \citep{dai-etal-2019-transformer} and
Sparse Attention \citep{sparseattn}.
Both of them performed attention in local windows of smaller sizes instead of across the whole sequence,
and differed in how they incorporated information from prior tokens.
This line of work was used in many models of note such as GPT3, and is commonly known as sliding window attention.
% aka sliding window aka in GPT3 and mistral
% https://arxiv.org/abs/1904.10509
This line of work has been further developed using methods like
Infini attention \citep{infiniattn}, which compresses prior tokens using a linear attention mechanism.

An alternative approach is to perform a lookup over prior tokens, such as in 
Landmark attention \citep{landmark}
Memorizing transformers \citep{wu2022memorizing}
or to do a lookup over an external datastore, which is commonly known as retrieval \citep{guu2020realm,atlas,retro}.
However, while these methods reduce computation, they do not reduce storage unless the KVs for lookup are offloaded from GPU memory.

\paragraph{Removing Softmax Attention}

Replacements to Softmax Attention are often referred to as SSMs or linear attention computations.
Regardless of name, most replace the attention component with an alternative which has constant space complexity during token generation w.r.t. the number of tokens generated.
This also reduces the time complexity of generation to be linear w.r.t. the number of tokens, instead of quadratic.
Various methods differ in how they parameterize their state.
\citet{linearattn,linformer}
proposed initial versions of linear attention.
Recent work has focused on using data dependent mechanisms to improve the state, such as in
GLA \citep{yang2024gated} Mamba\citep{gu2023mamba}
and RWKV v6 \citep{peng2024eagle}.

\paragraph{Groupings of Attention}
Most related are methods that use softmax attention but attempt to use a single KV pair for multiple queries.
GQA \citep{ainslie-etal-2023-gqa} and MQA \citep{shazeer2019mqa} do this by grouping keys and values across heads.

Concurrent work tries other strategies to share values between layers.
Deepseek-V2 \citep{deepseekai2024deepseekv2} proposes Multi Latent Attention, which uses a low rank projection of the keys and values.
You Only Cache Once \citep{sun2024cache} splits the model into two halves. The first half performs local attention and then generates a set of keys and values that are used for global attention across all layers in the second half.

\section{Conclusion}

Cross-Layer Attention is an effective method for reducing the KV cache memory storage footprint of transformer models by a factor of $2 \times$ with roughly equal perplexity.
Based on extensive experimental evaluation against well-tuned baselines at both the 1B- and 3B-parameter scales, we find that CLA advances the Pareto frontier for memory-efficient transformers.

%%%%%%%%%%%%%%%%%%%%%%%%%%%%%%%%%%%%%%%%%%%%%%%%%%%%%%%%%%%%

% TODO: Is this the best style to use? It wasn't included in the template.
\bibliographystyle{abbrvnat}
\bibliography{references}

\begin{thebibliography}{49}
\providecommand{\natexlab}[1]{#1}
\providecommand{\url}[1]{\texttt{#1}}
\expandafter\ifx\csname urlstyle\endcsname\relax
  \providecommand{\doi}[1]{doi: #1}\else
  \providecommand{\doi}{doi: \begingroup \urlstyle{rm}\Url}\fi

\bibitem[Ainslie et~al.(2023{\natexlab{a}})Ainslie, Lee-Thorp, de~Jong, Zemlyanskiy, Lebron, and Sanghai]{ainslie-etal-2023-gqa}
J.~Ainslie, J.~Lee-Thorp, M.~de~Jong, Y.~Zemlyanskiy, F.~Lebron, and S.~Sanghai.
\newblock {GQA}: Training generalized multi-query transformer models from multi-head checkpoints.
\newblock In H.~Bouamor, J.~Pino, and K.~Bali, editors, \emph{Proceedings of the 2023 Conference on Empirical Methods in Natural Language Processing}, pages 4895--4901, Singapore, Dec. 2023{\natexlab{a}}. Association for Computational Linguistics.
\newblock \doi{10.18653/v1/2023.emnlp-main.298}.
\newblock URL \url{https://aclanthology.org/2023.emnlp-main.298}.

\bibitem[Ainslie et~al.(2023{\natexlab{b}})Ainslie, Lee-Thorp, de~Jong, Zemlyanskiy, Lebrón, and Sanghai]{ainslie2023gqa}
J.~Ainslie, J.~Lee-Thorp, M.~de~Jong, Y.~Zemlyanskiy, F.~Lebrón, and S.~Sanghai.
\newblock Gqa: Training generalized multi-query transformer models from multi-head checkpoints, 2023{\natexlab{b}}.

\bibitem[Ansel et~al.(2024)Ansel, Yang, He, Gimelshein, Jain, Voznesensky, Bao, Bell, Berard, Burovski, Chauhan, Chourdia, Constable, Desmaison, DeVito, Ellison, Feng, Gong, Gschwind, Hirsh, Huang, Kalambarkar, Kirsch, Lazos, Lezcano, Liang, Liang, Lu, Luk, Maher, Pan, Puhrsch, Reso, Saroufim, Siraichi, Suk, Suo, Tillet, Wang, Wang, Wen, Zhang, Zhao, Zhou, Zou, Mathews, Chanan, Wu, and Chintala]{Ansel_PyTorch_2_Faster_2024}
J.~Ansel, E.~Yang, H.~He, N.~Gimelshein, A.~Jain, M.~Voznesensky, B.~Bao, P.~Bell, D.~Berard, E.~Burovski, G.~Chauhan, A.~Chourdia, W.~Constable, A.~Desmaison, Z.~DeVito, E.~Ellison, W.~Feng, J.~Gong, M.~Gschwind, B.~Hirsh, S.~Huang, K.~Kalambarkar, L.~Kirsch, M.~Lazos, M.~Lezcano, Y.~Liang, J.~Liang, Y.~Lu, C.~Luk, B.~Maher, Y.~Pan, C.~Puhrsch, M.~Reso, M.~Saroufim, M.~Y. Siraichi, H.~Suk, M.~Suo, P.~Tillet, E.~Wang, X.~Wang, W.~Wen, S.~Zhang, X.~Zhao, K.~Zhou, R.~Zou, A.~Mathews, G.~Chanan, P.~Wu, and S.~Chintala.
\newblock {PyTorch 2: Faster Machine Learning Through Dynamic Python Bytecode Transformation and Graph Compilation}.
\newblock In \emph{29th ACM International Conference on Architectural Support for Programming Languages and Operating Systems, Volume 2 (ASPLOS '24)}. ACM, Apr. 2024.
\newblock \doi{10.1145/3620665.3640366}.
\newblock URL \url{https://pytorch.org/assets/pytorch2-2.pdf}.

\bibitem[Black et~al.(2022)Black, Biderman, Hallahan, Anthony, Gao, Golding, He, Leahy, McDonell, Phang, Pieler, Prashanth, Purohit, Reynolds, Tow, Wang, and Weinbach]{black2022gptneox20b}
S.~Black, S.~Biderman, E.~Hallahan, Q.~Anthony, L.~Gao, L.~Golding, H.~He, C.~Leahy, K.~McDonell, J.~Phang, M.~Pieler, U.~S. Prashanth, S.~Purohit, L.~Reynolds, J.~Tow, B.~Wang, and S.~Weinbach.
\newblock Gpt-neox-20b: An open-source autoregressive language model, 2022.

\bibitem[Borgeaud et~al.(2022)Borgeaud, Mensch, Hoffmann, Cai, Rutherford, Millican, Van Den~Driessche, Lespiau, Damoc, Clark, De~Las~Casas, Guy, Menick, Ring, Hennigan, Huang, Maggiore, Jones, Cassirer, Brock, Paganini, Irving, Vinyals, Osindero, Simonyan, Rae, Elsen, and Sifre]{retro}
S.~Borgeaud, A.~Mensch, J.~Hoffmann, T.~Cai, E.~Rutherford, K.~Millican, G.~B. Van Den~Driessche, J.-B. Lespiau, B.~Damoc, A.~Clark, D.~De~Las~Casas, A.~Guy, J.~Menick, R.~Ring, T.~Hennigan, S.~Huang, L.~Maggiore, C.~Jones, A.~Cassirer, A.~Brock, M.~Paganini, G.~Irving, O.~Vinyals, S.~Osindero, K.~Simonyan, J.~Rae, E.~Elsen, and L.~Sifre.
\newblock Improving language models by retrieving from trillions of tokens.
\newblock In K.~Chaudhuri, S.~Jegelka, L.~Song, C.~Szepesvari, G.~Niu, and S.~Sabato, editors, \emph{Proceedings of the 39th International Conference on Machine Learning}, volume 162 of \emph{Proceedings of Machine Learning Research}, pages 2206--2240. PMLR, 17--23 Jul 2022.
\newblock URL \url{https://proceedings.mlr.press/v162/borgeaud22a.html}.

\bibitem[Child et~al.(2019)Child, Gray, Radford, and Sutskever]{sparseattn}
R.~Child, S.~Gray, A.~Radford, and I.~Sutskever.
\newblock Generating long sequences with sparse transformers.
\newblock \emph{CoRR}, abs/1904.10509, 2019.
\newblock URL \url{http://arxiv.org/abs/1904.10509}.

\bibitem[Chowdhery et~al.(2022)Chowdhery, Narang, Devlin, Bosma, Mishra, Roberts, Barham, Chung, Sutton, Gehrmann, Schuh, Shi, Tsvyashchenko, Maynez, Rao, Barnes, Tay, Shazeer, Prabhakaran, Reif, Du, Hutchinson, Pope, Bradbury, Austin, Isard, Gur-Ari, Yin, Duke, Levskaya, Ghemawat, Dev, Michalewski, Garcia, Misra, Robinson, Fedus, Zhou, Ippolito, Luan, Lim, Zoph, Spiridonov, Sepassi, Dohan, Agrawal, Omernick, Dai, Pillai, Pellat, Lewkowycz, Moreira, Child, Polozov, Lee, Zhou, Wang, Saeta, Diaz, Firat, Catasta, Wei, Meier-Hellstern, Eck, Dean, Petrov, and Fiedel]{chowdhery2022palm}
A.~Chowdhery, S.~Narang, J.~Devlin, M.~Bosma, G.~Mishra, A.~Roberts, P.~Barham, H.~W. Chung, C.~Sutton, S.~Gehrmann, P.~Schuh, K.~Shi, S.~Tsvyashchenko, J.~Maynez, A.~Rao, P.~Barnes, Y.~Tay, N.~Shazeer, V.~Prabhakaran, E.~Reif, N.~Du, B.~Hutchinson, R.~Pope, J.~Bradbury, J.~Austin, M.~Isard, G.~Gur-Ari, P.~Yin, T.~Duke, A.~Levskaya, S.~Ghemawat, S.~Dev, H.~Michalewski, X.~Garcia, V.~Misra, K.~Robinson, L.~Fedus, D.~Zhou, D.~Ippolito, D.~Luan, H.~Lim, B.~Zoph, A.~Spiridonov, R.~Sepassi, D.~Dohan, S.~Agrawal, M.~Omernick, A.~M. Dai, T.~S. Pillai, M.~Pellat, A.~Lewkowycz, E.~Moreira, R.~Child, O.~Polozov, K.~Lee, Z.~Zhou, X.~Wang, B.~Saeta, M.~Diaz, O.~Firat, M.~Catasta, J.~Wei, K.~Meier-Hellstern, D.~Eck, J.~Dean, S.~Petrov, and N.~Fiedel.
\newblock Palm: Scaling language modeling with pathways, 2022.

\bibitem[Dai et~al.(2019)Dai, Yang, Yang, Carbonell, Le, and Salakhutdinov]{dai-etal-2019-transformer}
Z.~Dai, Z.~Yang, Y.~Yang, J.~Carbonell, Q.~Le, and R.~Salakhutdinov.
\newblock Transformer-{XL}: Attentive language models beyond a fixed-length context.
\newblock In A.~Korhonen, D.~Traum, and L.~M{\`a}rquez, editors, \emph{Proceedings of the 57th Annual Meeting of the Association for Computational Linguistics}, pages 2978--2988, Florence, Italy, July 2019. Association for Computational Linguistics.
\newblock \doi{10.18653/v1/P19-1285}.
\newblock URL \url{https://aclanthology.org/P19-1285}.

\bibitem[Dao(2023)]{dao2023flashattention2}
T.~Dao.
\newblock Flashattention-2: Faster attention with better parallelism and work partitioning, 2023.

\bibitem[Dao et~al.(2022)Dao, Fu, Ermon, Rudra, and R{\'e}]{dao2022flashattention}
T.~Dao, D.~Y. Fu, S.~Ermon, A.~Rudra, and C.~R{\'e}.
\newblock Flash{A}ttention: Fast and memory-efficient exact attention with {IO}-awareness.
\newblock In \emph{Advances in Neural Information Processing Systems}, 2022.

\bibitem[DeepSeek-AI(2024)]{deepseekai2024deepseekv2}
DeepSeek-AI.
\newblock Deepseek-v2: A strong, economical, and efficient mixture-of-experts language model, 2024.

\bibitem[Gao et~al.(2024)Gao, He, Sharma, Kang, Jevdjic, Deng, Yang, Yu, and Zuo]{gao2024attentionstore}
B.~Gao, Z.~He, P.~Sharma, Q.~Kang, D.~Jevdjic, J.~Deng, X.~Yang, Z.~Yu, and P.~Zuo.
\newblock Attentionstore: Cost-effective attention reuse across multi-turn conversations in large language model serving, 2024.

\bibitem[Gao et~al.(2023)Gao, Tow, Abbasi, Biderman, Black, DiPofi, Foster, Golding, Hsu, Le~Noac'h, Li, McDonell, Muennighoff, Ociepa, Phang, Reynolds, Schoelkopf, Skowron, Sutawika, Tang, Thite, Wang, Wang, and Zou]{lm-eval-harness}
L.~Gao, J.~Tow, B.~Abbasi, S.~Biderman, S.~Black, A.~DiPofi, C.~Foster, L.~Golding, J.~Hsu, A.~Le~Noac'h, H.~Li, K.~McDonell, N.~Muennighoff, C.~Ociepa, J.~Phang, L.~Reynolds, H.~Schoelkopf, A.~Skowron, L.~Sutawika, E.~Tang, A.~Thite, B.~Wang, K.~Wang, and A.~Zou.
\newblock A framework for few-shot language model evaluation, 12 2023.
\newblock URL \url{https://zenodo.org/records/10256836}.

\bibitem[Ge et~al.(2024)Ge, Zhang, Liu, Zhang, Han, and Gao]{ge2024fastgen}
S.~Ge, Y.~Zhang, L.~Liu, M.~Zhang, J.~Han, and J.~Gao.
\newblock Model tells you what to discard: Adaptive kv cache compression for llms, 2024.

\bibitem[{Google}(2024)]{googleai2024context}
{Google}.
\newblock Context caching guide.
\newblock \url{https://ai.google.dev/gemini-api/docs/caching}, 2024.
\newblock Accessed: 2024-05-20.

\bibitem[Gu and Dao(2023)]{gu2023mamba}
A.~Gu and T.~Dao.
\newblock Mamba: Linear-time sequence modeling with selective state spaces, 2023.

\bibitem[Guu et~al.(2020)Guu, Lee, Tung, Pasupat, and Chang]{guu2020realm}
K.~Guu, K.~Lee, Z.~Tung, P.~Pasupat, and M.-W. Chang.
\newblock Realm: retrieval-augmented language model pre-training.
\newblock In \emph{Proceedings of the 37th International Conference on Machine Learning}, ICML'20. JMLR.org, 2020.

\bibitem[Hooper et~al.(2024)Hooper, Kim, Mohammadzadeh, Mahoney, Shao, Keutzer, and Gholami]{hooper2024kvquant}
C.~Hooper, S.~Kim, H.~Mohammadzadeh, M.~W. Mahoney, Y.~S. Shao, K.~Keutzer, and A.~Gholami.
\newblock Kvquant: Towards 10 million context length llm inference with kv cache quantization, 2024.

\bibitem[Huang et~al.(2019)Huang, Cheng, Bapna, Firat, Chen, Chen, Lee, Ngiam, Le, Wu, and Chen]{10.5555/3454287.3454297}
Y.~Huang, Y.~Cheng, A.~Bapna, O.~Firat, M.~X. Chen, D.~Chen, H.~Lee, J.~Ngiam, Q.~V. Le, Y.~Wu, and Z.~Chen.
\newblock \emph{GPipe: efficient training of giant neural networks using pipeline parallelism}.
\newblock Curran Associates Inc., Red Hook, NY, USA, 2019.

\bibitem[Izacard et~al.(2024)Izacard, Lewis, Lomeli, Hosseini, Petroni, Schick, Dwivedi-Yu, Joulin, Riedel, and Grave]{atlas}
G.~Izacard, P.~Lewis, M.~Lomeli, L.~Hosseini, F.~Petroni, T.~Schick, J.~Dwivedi-Yu, A.~Joulin, S.~Riedel, and E.~Grave.
\newblock Atlas: few-shot learning with retrieval augmented language models.
\newblock \emph{J. Mach. Learn. Res.}, 24\penalty0 (1), mar 2024.
\newblock ISSN 1532-4435.

\bibitem[Kalamkar et~al.(2019)Kalamkar, Mudigere, Mellempudi, Das, Banerjee, Avancha, Vooturi, Jammalamadaka, Huang, Yuen, Yang, Park, Heinecke, Georganas, Srinivasan, Kundu, Smelyanskiy, Kaul, and Dubey]{kalamkar2019study}
D.~Kalamkar, D.~Mudigere, N.~Mellempudi, D.~Das, K.~Banerjee, S.~Avancha, D.~T. Vooturi, N.~Jammalamadaka, J.~Huang, H.~Yuen, J.~Yang, J.~Park, A.~Heinecke, E.~Georganas, S.~Srinivasan, A.~Kundu, M.~Smelyanskiy, B.~Kaul, and P.~Dubey.
\newblock A study of bfloat16 for deep learning training, 2019.

\bibitem[Katharopoulos et~al.(2020)Katharopoulos, Vyas, Pappas, and Fleuret]{linearattn}
A.~Katharopoulos, A.~Vyas, N.~Pappas, and F.~Fleuret.
\newblock Transformers are rnns: fast autoregressive transformers with linear attention.
\newblock In \emph{Proceedings of the 37th International Conference on Machine Learning}, ICML'20. JMLR.org, 2020.

\bibitem[Liu et~al.(2024)Liu, Li, Cheng, Ray, Huang, Zhang, Du, Yao, Lu, Ananthanarayanan, Maire, Hoffmann, Holtzman, and Jiang]{liu2024cachegen}
Y.~Liu, H.~Li, Y.~Cheng, S.~Ray, Y.~Huang, Q.~Zhang, K.~Du, J.~Yao, S.~Lu, G.~Ananthanarayanan, M.~Maire, H.~Hoffmann, A.~Holtzman, and J.~Jiang.
\newblock Cachegen: Kv cache compression and streaming for fast language model serving, 2024.

\bibitem[Liu et~al.(2023)Liu, Desai, Liao, Wang, Xie, Xu, Kyrillidis, and Shrivastava]{liu2023scissorhands}
Z.~Liu, A.~Desai, F.~Liao, W.~Wang, V.~Xie, Z.~Xu, A.~Kyrillidis, and A.~Shrivastava.
\newblock Scissorhands: Exploiting the persistence of importance hypothesis for llm kv cache compression at test time, 2023.

\bibitem[Loshchilov and Hutter(2017)]{loshchilov2017sgdr}
I.~Loshchilov and F.~Hutter.
\newblock {SGDR}: Stochastic gradient descent with warm restarts.
\newblock In \emph{International Conference on Learning Representations}, 2017.
\newblock URL \url{https://openreview.net/forum?id=Skq89Scxx}.

\bibitem[Loshchilov and Hutter(2019)]{loshchilov2018decoupled}
I.~Loshchilov and F.~Hutter.
\newblock Decoupled weight decay regularization.
\newblock In \emph{International Conference on Learning Representations}, 2019.
\newblock URL \url{https://openreview.net/forum?id=Bkg6RiCqY7}.

\bibitem[Merity et~al.(2016)Merity, Xiong, Bradbury, and Socher]{merity2016pointer}
S.~Merity, C.~Xiong, J.~Bradbury, and R.~Socher.
\newblock Pointer sentinel mixture models, 2016.

\bibitem[Micikevicius et~al.(2018)Micikevicius, Narang, Alben, Diamos, Elsen, Garcia, Ginsburg, Houston, Kuchaiev, Venkatesh, and Wu]{micikevicius2018mixed}
P.~Micikevicius, S.~Narang, J.~Alben, G.~Diamos, E.~Elsen, D.~Garcia, B.~Ginsburg, M.~Houston, O.~Kuchaiev, G.~Venkatesh, and H.~Wu.
\newblock Mixed precision training, 2018.

\bibitem[Mohtashami and Jaggi(2023)]{landmark}
A.~Mohtashami and M.~Jaggi.
\newblock Random-access infinite context length for transformers.
\newblock In A.~Oh, T.~Naumann, A.~Globerson, K.~Saenko, M.~Hardt, and S.~Levine, editors, \emph{Advances in Neural Information Processing Systems}, volume~36, pages 54567--54585. Curran Associates, Inc., 2023.
\newblock URL \url{https://proceedings.neurips.cc/paper_files/paper/2023/file/ab05dc8bf36a9f66edbff6992ec86f56-Paper-Conference.pdf}.

\bibitem[Munkhdalai et~al.(2024)Munkhdalai, Faruqui, and Gopal]{infiniattn}
T.~Munkhdalai, M.~Faruqui, and S.~Gopal.
\newblock Leave no context behind: Efficient infinite context transformers with infini-attention, 2024.

\bibitem[Paszke et~al.(2019)Paszke, Gross, Massa, Lerer, Bradbury, Chanan, Killeen, Lin, Gimelshein, Antiga, Desmaison, Köpf, Yang, DeVito, Raison, Tejani, Chilamkurthy, Steiner, Fang, Bai, and Chintala]{paszke2019pytorch}
A.~Paszke, S.~Gross, F.~Massa, A.~Lerer, J.~Bradbury, G.~Chanan, T.~Killeen, Z.~Lin, N.~Gimelshein, L.~Antiga, A.~Desmaison, A.~Köpf, E.~Yang, Z.~DeVito, M.~Raison, A.~Tejani, S.~Chilamkurthy, B.~Steiner, L.~Fang, J.~Bai, and S.~Chintala.
\newblock Pytorch: An imperative style, high-performance deep learning library, 2019.

\bibitem[Peng et~al.(2024)Peng, Goldstein, Anthony, Albalak, Alcaide, Biderman, Cheah, Du, Ferdinan, Hou, Kazienko, GV, Kocoń, Koptyra, Krishna, au2, Muennighoff, Obeid, Saito, Song, Tu, Woźniak, Zhang, Zhao, Zhao, Zhou, Zhu, and Zhu]{peng2024eagle}
B.~Peng, D.~Goldstein, Q.~Anthony, A.~Albalak, E.~Alcaide, S.~Biderman, E.~Cheah, X.~Du, T.~Ferdinan, H.~Hou, P.~Kazienko, K.~K. GV, J.~Kocoń, B.~Koptyra, S.~Krishna, R.~M.~J. au2, N.~Muennighoff, F.~Obeid, A.~Saito, G.~Song, H.~Tu, S.~Woźniak, R.~Zhang, B.~Zhao, Q.~Zhao, P.~Zhou, J.~Zhu, and R.-J. Zhu.
\newblock Eagle and finch: Rwkv with matrix-valued states and dynamic recurrence, 2024.

\bibitem[Ramachandran et~al.(2017)Ramachandran, Zoph, and Le]{ramachandran2017searching}
P.~Ramachandran, B.~Zoph, and Q.~V. Le.
\newblock Searching for activation functions, 2017.

\bibitem[Shazeer(2019)]{shazeer2019mqa}
N.~Shazeer.
\newblock Fast transformer decoding: One write-head is all you need, 2019.

\bibitem[Shazeer(2020)]{shazeer2020glu}
N.~Shazeer.
\newblock Glu variants improve transformer, 2020.

\bibitem[Sheng et~al.(2023)Sheng, Zheng, Yuan, Li, Ryabinin, Chen, Liang, R\'{e}, Stoica, and Zhang]{flexgen}
Y.~Sheng, L.~Zheng, B.~Yuan, Z.~Li, M.~Ryabinin, B.~Chen, P.~Liang, C.~R\'{e}, I.~Stoica, and C.~Zhang.
\newblock Flexgen: high-throughput generative inference of large language models with a single gpu.
\newblock In \emph{Proceedings of the 40th International Conference on Machine Learning}, ICML'23. JMLR.org, 2023.

\bibitem[Shoeybi et~al.(2020)Shoeybi, Patwary, Puri, LeGresley, Casper, and Catanzaro]{shoeybi2020megatronlm}
M.~Shoeybi, M.~Patwary, R.~Puri, P.~LeGresley, J.~Casper, and B.~Catanzaro.
\newblock Megatron-lm: Training multi-billion parameter language models using model parallelism, 2020.

\bibitem[Soboleva et~al.(2023)Soboleva, Al-Khateeb, Myers, Steeves, Hestness, and Dey]{cerebras2023slimpajama}
D.~Soboleva, F.~Al-Khateeb, R.~Myers, J.~R. Steeves, J.~Hestness, and N.~Dey.
\newblock {SlimPajama: A 627B token cleaned and deduplicated version of RedPajama}.
\newblock \url{https://www.cerebras.net/blog/slimpajama-a-627b-token-cleaned-and-deduplicated-version-of-redpajama}, 2023.
\newblock URL \url{https://huggingface.co/datasets/cerebras/SlimPajama-627B}.

\bibitem[Su et~al.(2023)Su, Lu, Pan, Murtadha, Wen, and Liu]{su2023roformer}
J.~Su, Y.~Lu, S.~Pan, A.~Murtadha, B.~Wen, and Y.~Liu.
\newblock Roformer: Enhanced transformer with rotary position embedding, 2023.

\bibitem[Sun et~al.(2024)Sun, Dong, Zhu, Huang, Wang, Ma, Zhang, Wang, and Wei]{sun2024cache}
Y.~Sun, L.~Dong, Y.~Zhu, S.~Huang, W.~Wang, S.~Ma, Q.~Zhang, J.~Wang, and F.~Wei.
\newblock You only cache once: Decoder-decoder architectures for language models, 2024.

\bibitem[Touvron et~al.(2023)Touvron, Martin, Stone, Albert, Almahairi, Babaei, Bashlykov, Batra, Bhargava, Bhosale, Bikel, Blecher, Ferrer, Chen, Cucurull, Esiobu, Fernandes, Fu, Fu, Fuller, Gao, Goswami, Goyal, Hartshorn, Hosseini, Hou, Inan, Kardas, Kerkez, Khabsa, Kloumann, Korenev, Koura, Lachaux, Lavril, Lee, Liskovich, Lu, Mao, Martinet, Mihaylov, Mishra, Molybog, Nie, Poulton, Reizenstein, Rungta, Saladi, Schelten, Silva, Smith, Subramanian, Tan, Tang, Taylor, Williams, Kuan, Xu, Yan, Zarov, Zhang, Fan, Kambadur, Narang, Rodriguez, Stojnic, Edunov, and Scialom]{touvron2023llama}
H.~Touvron, L.~Martin, K.~Stone, P.~Albert, A.~Almahairi, Y.~Babaei, N.~Bashlykov, S.~Batra, P.~Bhargava, S.~Bhosale, D.~Bikel, L.~Blecher, C.~C. Ferrer, M.~Chen, G.~Cucurull, D.~Esiobu, J.~Fernandes, J.~Fu, W.~Fu, B.~Fuller, C.~Gao, V.~Goswami, N.~Goyal, A.~Hartshorn, S.~Hosseini, R.~Hou, H.~Inan, M.~Kardas, V.~Kerkez, M.~Khabsa, I.~Kloumann, A.~Korenev, P.~S. Koura, M.-A. Lachaux, T.~Lavril, J.~Lee, D.~Liskovich, Y.~Lu, Y.~Mao, X.~Martinet, T.~Mihaylov, P.~Mishra, I.~Molybog, Y.~Nie, A.~Poulton, J.~Reizenstein, R.~Rungta, K.~Saladi, A.~Schelten, R.~Silva, E.~M. Smith, R.~Subramanian, X.~E. Tan, B.~Tang, R.~Taylor, A.~Williams, J.~X. Kuan, P.~Xu, Z.~Yan, I.~Zarov, Y.~Zhang, A.~Fan, M.~Kambadur, S.~Narang, A.~Rodriguez, R.~Stojnic, S.~Edunov, and T.~Scialom.
\newblock Llama 2: Open foundation and fine-tuned chat models, 2023.

\bibitem[Vaswani et~al.(2017)Vaswani, Shazeer, Parmar, Uszkoreit, Jones, Gomez, Kaiser, and Polosukhin]{Vaswani+2017}
A.~Vaswani, N.~Shazeer, N.~Parmar, J.~Uszkoreit, L.~Jones, A.~N. Gomez, L.~u. Kaiser, and I.~Polosukhin.
\newblock Attention is all you need.
\newblock In \emph{Advances in Neural Information Processing Systems}, volume~30. Curran Associates, Inc., 2017.
\newblock URL \url{https://proceedings.neurips.cc/paper_files/paper/2017/file/3f5ee243547dee91fbd053c1c4a845aa-Paper.pdf}.

\bibitem[Wang et~al.(2019)Wang, Cho, and Gu]{wang2019neural}
C.~Wang, K.~Cho, and J.~Gu.
\newblock Neural machine translation with byte-level subwords, 2019.

\bibitem[Wang et~al.(2020)Wang, Li, Khabsa, Fang, and Ma]{linformer}
S.~Wang, B.~Z. Li, M.~Khabsa, H.~Fang, and H.~Ma.
\newblock Linformer: Self-attention with linear complexity.
\newblock \emph{CoRR}, abs/2006.04768, 2020.
\newblock URL \url{https://arxiv.org/abs/2006.04768}.

\bibitem[Wu et~al.(2022)Wu, Rabe, Hutchins, and Szegedy]{wu2022memorizing}
Y.~Wu, M.~N. Rabe, D.~Hutchins, and C.~Szegedy.
\newblock Memorizing transformers.
\newblock In \emph{International Conference on Learning Representations}, 2022.
\newblock URL \url{https://openreview.net/forum?id=TrjbxzRcnf-}.

\bibitem[Yang et~al.(2024)Yang, Wang, Shen, Panda, and Kim]{yang2024gated}
S.~Yang, B.~Wang, Y.~Shen, R.~Panda, and Y.~Kim.
\newblock Gated linear attention transformers with hardware-efficient training, 2024.

\bibitem[Zhang et~al.(2024)Zhang, Yi, Xu, and Shrivastava]{zhang2024kvonebit}
T.~Zhang, J.~Yi, Z.~Xu, and A.~Shrivastava.
\newblock Kv cache is 1 bit per channel: Efficient large language model inference with coupled quantization, 2024.

\bibitem[Zhang et~al.(2023)Zhang, Sheng, Zhou, Chen, Zheng, Cai, Song, Tian, R\'{e}, Barrett, Wang, and Chen]{h2o}
Z.~Zhang, Y.~Sheng, T.~Zhou, T.~Chen, L.~Zheng, R.~Cai, Z.~Song, Y.~Tian, C.~R\'{e}, C.~Barrett, Z.~A. Wang, and B.~Chen.
\newblock H2o: Heavy-hitter oracle for efficient generative inference of large language models.
\newblock In A.~Oh, T.~Naumann, A.~Globerson, K.~Saenko, M.~Hardt, and S.~Levine, editors, \emph{Advances in Neural Information Processing Systems}, volume~36, pages 34661--34710. Curran Associates, Inc., 2023.
\newblock URL \url{https://proceedings.neurips.cc/paper_files/paper/2023/file/6ceefa7b15572587b78ecfcebb2827f8-Paper-Conference.pdf}.

\bibitem[Zhu et~al.(2023)Zhu, Li, Liu, Ma, and Wang]{zhu2023survey}
X.~Zhu, J.~Li, Y.~Liu, C.~Ma, and W.~Wang.
\newblock A survey on model compression for large language models, 2023.

\end{thebibliography}

\appendix

\section{Learning rate sweeps}\label{sec:lr-sweep}

Here we present the results of our learning rate sweeps at the 1B and 3B parameter scales:

\begin{figure}[h]
    \centering
    \includegraphics[scale=0.85]{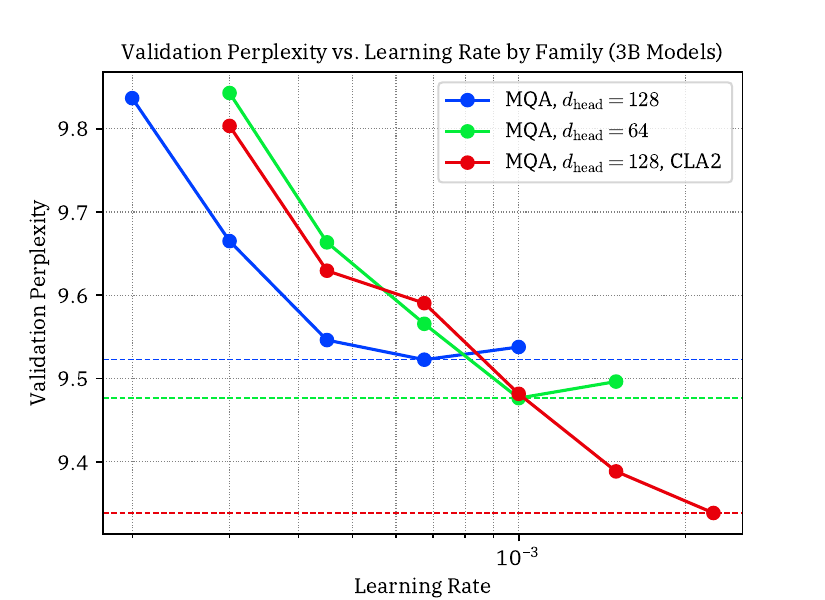}
    \caption{3B Learning Rate Sweep}
    \label{fig:3B-lr}
\end{figure}
\begin{figure}[h]
    \centering
    \includegraphics[scale=0.85]{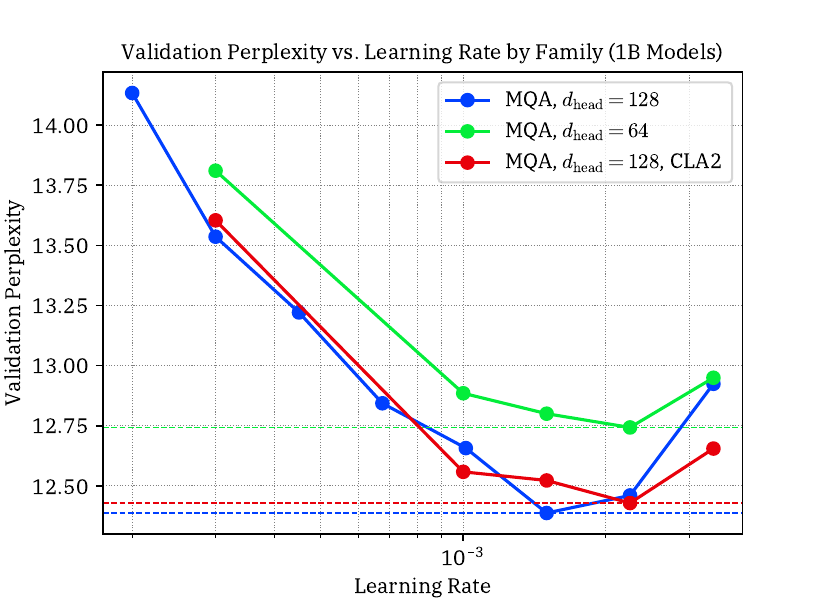}
    \caption{1B Learning Rate Sweep}
    \label{fig:1B-lr}
\end{figure}

\end{document}